\documentclass[a4paper, 10pt, conference]{ieeeconf}      
\usepackage{FG2021}

\FGfinalcopy 

\IEEEoverridecommandlockouts                              
\usepackage{times}
\usepackage{epsfig}
\usepackage{graphicx}
\usepackage{amsmath}
\usepackage{amssymb}

\usepackage{multirow}
\usepackage{color}
\usepackage{balance}
\usepackage{url}
\usepackage[pagebackref=true,breaklinks=true,colorlinks,bookmarks=false]{hyperref}
\usepackage[stable]{footmisc}

\newcommand{\so}[1]{{\color{black} #1}}

\def\FGPaperID{62} 

\title{\LARGE \bf
Automated Detection of Equine Facial Action Units}

\author{Zhenghong Li$^{1,2}$ ~~~~ 
Sofia Broomé$^1$ ~~~~ Pia Haubro Andersen$^4$ ~~~~ Hedvig Kjellström$^{1,3}$\\
$^1$ KTH Royal Institute of Technology, Sweden {\tt sbroome,hedvig@kth.se}\\
$^2$ Stony Brook University, USA {\tt zhenghong.li@stonybrook.edu} ~~~~ $^3$ Silo AI, Sweden\\
$^4$ Swedish University of Agricultural Sciences, Sweden {\tt pia.haubro.andersen@slu.se}}

\begin{document}

\ifFGfinal
\thispagestyle{empty}
\pagestyle{empty}
\else
\author{Anonymous FG2021 submission\\ Paper ID \FGPaperID \\ \\ \\}
\pagestyle{plain}
\fi
\maketitle

\begin{abstract}
The recently developed Equine Facial Coding System (EquiFACS) provides a precise and exhaustive, but laborious, manual labelling method of facial action units of the horse. To automate parts of this process, we propose a Deep Learning-based method to detect EquiFACS units automatically from images. We use a cascade framework; we firstly train several object detectors to detect the predefined Region-of-Interest (ROI), and secondly apply binary classifiers for each action unit in related regions. We experiment with both regular CNNs and a more tailored model transferred from human facial action unit recognition. Promising initial results are presented for nine action units in the eye and lower face regions. Code for the project is publicly available. 
\end{abstract}



\section{Introduction}
The horse is a highly social species, and facial communication is of utmost importance for the function of the herd. 
In accordance, the horse has a remarkable repertoire of facial expressions which may be described by 17 degrees of freedom, so called actions units \cite{wathan2015equifacs}. This repertoire is smaller than for humans which have 27 action units \cite{ekman1978facial}, but larger than for example the chimpanzee repertoire of 13 action units \cite{diogo2009face}.

While the detailed analysis of the facial expressions of people to assess their emotions is mature \cite{ekman1993emotion}, almost nothing is known about the association between facial activity and emotional states of animals. This is primarily due to the lack of self-report of emotions and other inner states in animals. Nevertheless, facial expressions are expected to convey important information of animal welfare \cite{descovich2017face}, but methodologies for investigations are lacking.

In the past few years, great progress has been made in the field of Computer Vision. With the adoption of Deep Learning models, such as Convolutional Neural Networks (CNN), in Computer Vision, in some tasks such as image classification, the accuracies of computer models are even competitive with human capabilities. Related works for human facial action unit detection have also made progress in these years.

Therefore, in this work, we investigate the possibility of
automatically
recognizing horse facial action units. We currently focus on how to 
do this from still images. Even if the facial configurations of horses and humans are very different, a remarkably high number of action units are conserved across species \cite{williams2002pain}. We therefore transfer methods for human action unit detection to horses.

There are two main contributions of our project:
\begin{itemize}
\setlength{\itemsep}{0pt}

    \item We propose a cascade framework for the recognition of horse facial action units.
    \item We apply standard models for general image classification as well as for human facial action unit recognition to horses within our framework and compare their performance across multiple experimental settings. 
\end{itemize}


\section{Related Work}

Facial expressions can be described as combinations of different facial action units. A facial action unit is based on the visible movement of a facial muscle lying under the skin \cite{shao2019facial}. 
In 1978, Ekman and Friesen proposed the Facial Action Coding System (FACS) \cite{ekman1978facial}. Through electrically stimulating individual muscles and learning to control them voluntarily, each action unit was associated with one or more facial muscles \cite{cohn2007observer}. The recording of facial actions is entirely atheoretical; any inference of their meaning takes places during the later analysis.  In 2002, Ekman et al.~\cite{ekman2002facial} proposed the final version of human FACS which since has been widely used for research in human emotion recognition. 

Inspired by the progress of human FACS, Wathan et al.~\cite{wathan2015equifacs} created EquiFACS. As for FACS, EquiFACS consists of action units (AUs) and action descriptors (ADs). In addition, the movements of the ears of horses are specifically named as ear action descriptors (EADs). Until recently, EquiFACS has not been used for research in animal emotions, due to the very time consuming manual labelling. An initial study of facial expressions of pain in horses \cite{rashid2020equifacs} showed that pain indeed is associated with increased frequencies of certain facial AUs. These AUs were anatomically located in the ear region, around the eyes, nostrils and muzzle. A major limitation of that study was the small sample size, which was attributed to the extremely resource demanding, but necessary, hand labelling of the videos. A prerequisite for more research in animal emotions using AUs is therefore development of methods that allow automated AU detection.

Pain recognition in animals via pre-defined facial features has previously been explored for sheep \cite{lu2017estimating, Pessanha2020} and for horses and donkeys \cite{hummel2020automatic}. Compared to our method, these works rely on more coarse-grained underlying facial expression representations, albeit using precise landmark extraction to extract regions of interest. 
The simpler structure increases robustness but limits the range and precision of expressions that can be represented.
A third approach is to learn the underlying representation of pain expressions in horses from raw data in an end-to-end manner \so{\cite{broome2019dynamics}}, without imposing any designed coding system. \so{In \cite{broome2019dynamics}, the authors}
used a recurrent neural network structure that exploited both temporal and spatial information from video of horses, and found that temporal information (video) is important for pain assessment. A future research direction is to study the interplay between data-driven learned emotion expression representations and EquiFACS.

\so{Ever since} Krizhevsky et al.~proposed AlexNet \cite{krizhevsky2012imagenet}, deep CNNs have been replacing the traditional methods in image classification fields with their outstanding performance. After AlexNet, deeper models such as VGG \cite{simonyan2014very} and ResNet \cite{he2016deep} have been proposed and applied as feature extractors in various fields. In our work, we
\so{chose} CNNs as the classifiers of AUs.

CNNs are also widely used for object detection. In this work, an object detector network is employed to detect pre-defined regions of interest (ROI). Object detectors can be divided into two categories: one-stage methods and two-stage methods. One-stage methods such as YOLOv3 \cite{Redmon2018YOLOv3} and SSD \cite{Liu_2016} generate anchor proposals and perform detection in one stage and can be trained in an end-to-end manner. Two-stage methods such as Faster-RCNN \cite{NIPS2015_5638} first generate anchor box proposals via a region proposal network and then use ROI-Pooling to crop the related features out for the final prediction. 

Previous works in human facial AU recognition from still images usually employ regional learning. Zhao et al.~\cite{zhao2016deep} inserted a region layer into a classical CNN to learn the features of AUs on sparse facial regions, and trained the model via multi-label learning. Shao et al.~\cite{shao2019facial} further cascaded region layers with different sizes of patches together and employed an attention mechanism for AU detection.


\section{Data}
In total, the dataset used for this study contains 20180 labeled video clips across 31 AUs or ADs, \so{with durations} ranging from 0.05 seconds to 2 minutes. The data is recorded across eight horse subjects. We randomly sample one frame from each labeled clip to use as input for our classifier.

 The class distribution is quite uneven. 
 There are, e.g., 5280 labeled samples
for EAD104 (ear rotator), but only one for AD160 (lower lip relax). For our experiments, we selected the 11 categories listed in Table \ref{table:au}. Each 
contains more than 200 labeled crops, which we consider to be the minimal sufficient number of samples 
for the training, validation and test sets. 
However, we quickly found that the ear action descriptors were not suited to detect using still images, since they are defined by movement. For this reason, we chose to exclude EAD101 and EAD104 from our experiments.


We perform subject-exclusive eight-fold validation across the different horses, using six for training, one for validation and one for testing in each fold.

As for the sampled images, the original sizes are $1910 \times 1080$ or $1272 \times 720$. For the face, eye and lower face crops, we first zero-pad the detected regions, to then resize them. Face crops are resized to $512 \times 512$, as they are then fed into YOLOv3-tiny whose default input size is $416 \times 416$. Eye and lower face crops are resized to $64 \times 64$ for the modified DRML and modified AlexNet classifier\so{s}, which can run on smaller input sizes.

\begin{table}[!t]
\centering
\caption{Selected Action Units (Action Descriptors).}
\vspace{-1.5mm}
\begin{tabular}{c|c|c}
\textbf{Code}  & \textbf{AU (AD)}       & \textbf{Labeled video clips} \\ \hline
AD1    & Eye white increase & 394    \\ \hline
AD19   & Tougue show        & 443    \\ \hline
AD38   & Nostril dilator    & 696    \\ \hline
AU101  & Inner brow raiser  & 1918   \\ \hline
AU145  & Blink              & 3856   \\ \hline
AU25   & Lips part          & 478    \\ \hline
AU47   & Half blink         & 1816   \\ \hline
AU5    & Upper lid raiser   & 208    \\ \hline
AUH13  & Nostril lift       & 353    \\ \hline
EAD101 & Ears forward       & 4778   \\ \hline
EAD104 & Ear rotator        & 5240   \\
\end{tabular}
\label{table:au}
\end{table}

\section{Methodology\footnote{Code for all methods presented in this section is publicly available in the repository \url{https://github.com/ZhenghLi/Automated-Detection-of-Equine-Facial-Action-Units}.}}

Considering the class imbalance, the dataset is not suited for multi-label classification. Initial tests were carried out in this fashion, but the model would get stuck in a local minimum where it predicted the dominant AUs to be true and the others to be false.
Therefore, we use multiple binary classifiers for the 
\so{nine} classes. For each binary classifier set-up, we randomly sample negative \so{samples}
to make the number of positive and negative \so{samples} equal. This was done for both the training, validation and test split of the data.

\begin{figure}[!t]
    \centering
    \includegraphics[width=0.475\textwidth]{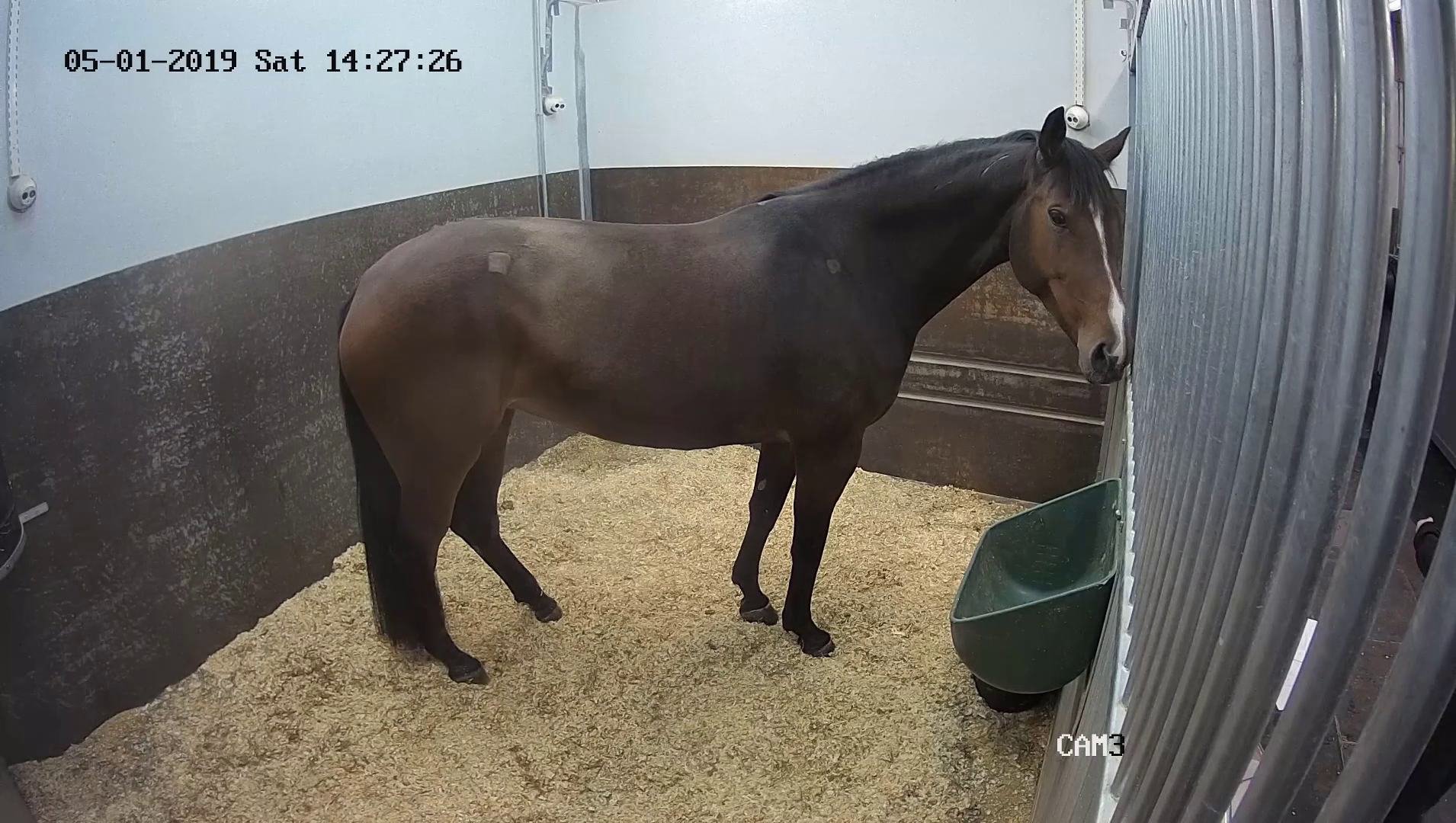}
        \vspace{-2.5mm}
    \caption{A raw example of AD1 eye white increase in our dataset}
    \label{fig:example}
\end{figure}

\begin{figure}[!t]
\vspace{3mm}
    \centering
    \includegraphics[width=0.475\textwidth]{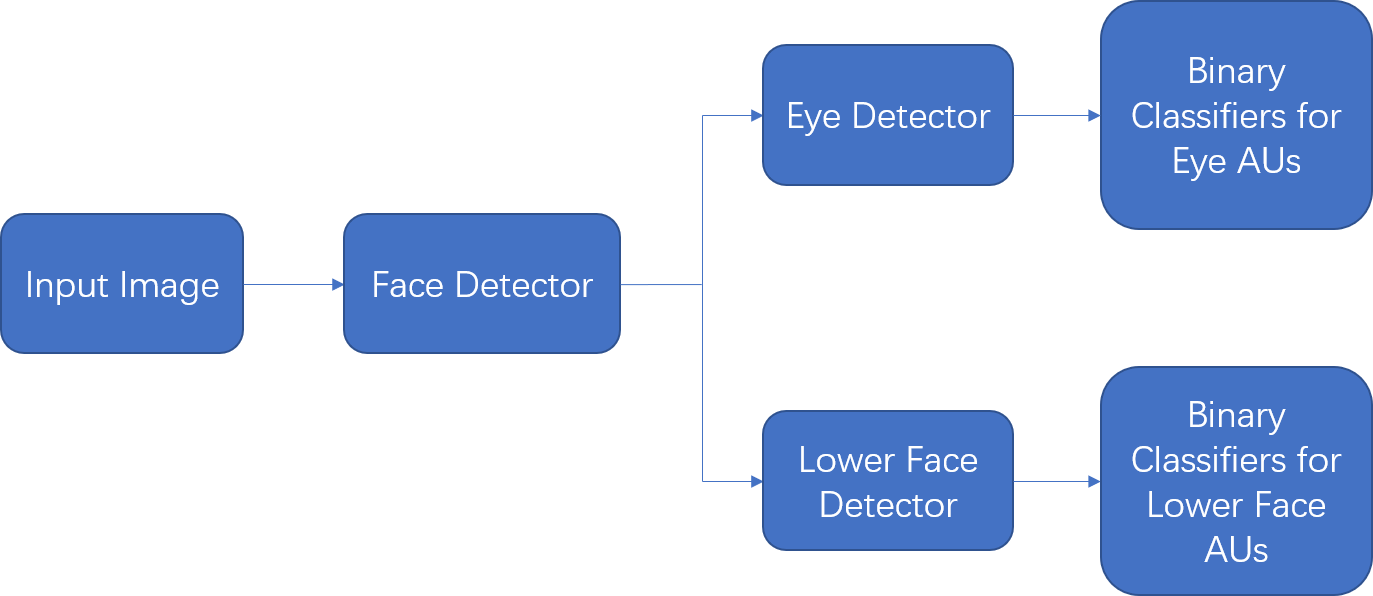}
        \vspace{-2.5mm}
    \caption{Our cascade framework for EquiFACS recognition. Note that each part is trained separately}
    \label{fig:framework}
\end{figure}

Further, binary classification for facial AUs is a highly fine-grained image classification task. As such, directly applying networks for common image classification tasks will fail to reach acceptable results. Noticing that the horse face is usually only a fraction of the raw frame (Fig. \ref{fig:example}), and inspired by the framework for sheep pain estimation by Lu et al.~\cite{lu2017estimating}, we propose our Deep Learning cascade framework (Fig.~\ref{fig:framework}) for horse facial AU recognition. For each input image, we first detect the horse face and cut it out. Then, depending on the facial location of the \so{action unit} class, we extract either the eye region or the lower face region (including nostrils and mouth) from the detected face region. This is because the eye regions and the lower face regions naturally take up even smaller fractions of the raw frames, and the detector network is not able to detect these regions directly. Finally,
\so{two CNN-based models}
for image classification
\so{are used} as binary classifiers for the respective classes belonging to these regions. Note that the classifiers are trained separately for each class.

\subsection{ROI Detector}
YOLOv3 \cite{Redmon2018YOLOv3} is a widely used object detector, with high performance with respect to both average precision and computation speed. For our task, we
\so{chose} one of its light-weight implementations, YOLOv3-tiny for the ROI detection, since it could be readily applied to our dataset with an acceptable performance.
\so{Knowing} that existing object detectors do not perform well on objects that are small relative to the frame,
we chose to cascade the detectors together to detect small regions. Specifically, we first trained a face detector, and then trained an eye region and a lower face detector, respectively, on the detected face crops.

\subsection{Classifier}
DRML \cite{zhao2016deep} is a classical model for human facial AU detection. We first experimented with directly training it on raw frames as well as on detected face regions.
 \so{Then, we applied it to detected eye or lower face regions, while replacing the region layer with a simple $3 \times 3$ convolution layer.}
In addition, we also applied the AlexNet \cite{krizhevsky2012imagenet} as
\so{a} classifier on all three levels of detail. When using DRML and AlexNet for crops of the eye or lower face regions (resizing the input to a resolution of $64 \times 64$), we modified the first convolutional 
\so{layer in each model} to use $5 \times 5$ convolutional kernels (instead of $11 \times 11$ as in the original models).


\section{Experimental Results}

\subsection{Model Exploration on AU101}
First, we explore which frameworks 
are suitable for horse facial AU recognition. We evaluate these on AU101 (inner brow raiser), because the key features of AU101 (the angular shape of the inner brows) are relatively easy to recognize in images and the class has more labeled \so{samples} than other relatively "easy" classes. Using eight-fold validation, we evaluate the performance of the DRML model and AlexNet on raw frames, detected face regions, and detected eye regions, in turn. Results are shown in Table \ref{table:au101result}.

\begin{table}[!t]
\centering
\caption{Binary classification results for AU101, preceded by region detection of different precision: whole frame, head region, or most precisely, eye region. Mean and standard deviation resulting from eight-fold cross-validation.}
\vspace{-1.5mm}
\scalebox{0.9}{
\begin{tabular}{|l|l|l|l|l|}
\hline
\multirow{2}{*}{Region} & \multicolumn{2}{c|}{DRML}             & \multicolumn{2}{c|}{AlexNet} \\ \cline{2-5} 
                        & Accuracy          & F1-score          & Accuracy     & F1-score      \\ \hline
Frame & 54.0$\pm$7.1          & 46.9$\pm$10.9         & 52.8$\pm$5.8     & 46.0$\pm$11.6     \\ \hline
Face                    & 53.7$\pm$6.0          & 47.6$\pm$12.9         & 53.6$\pm$4.3     & 51.0$\pm$13.5     \\ \hline
Eye                     & \textbf{58.1$\pm$4.8} & \textbf{60.7$\pm$6.9} & 57.0$\pm$6.5     & 58.0$\pm$10.7     \\ \hline
\end{tabular}}
\label{table:au101result}
\end{table}

\begin{figure*}[!t]
    \centering
    ~\hspace{-7.6mm}\includegraphics[width=1.005\textwidth]{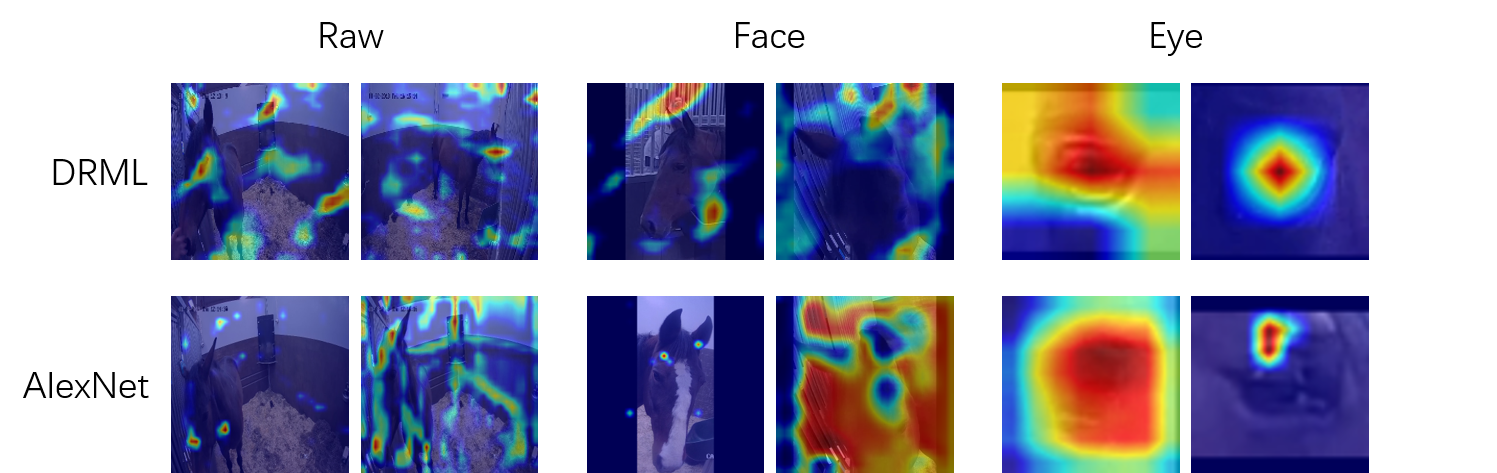}
    \vspace{-2.5mm}
    \caption{Grad-CAM saliency maps of the models for binary classification of AU101}
    \label{fig:au101result}
\end{figure*}

For both DRML and AlexNet, we observe that there is no large difference between classification on the raw frames and on face crops. The results are merely random. We further employed Grad-CAM \cite{Selvaraju_2019} to visualize what the critical portions of the images were for these classifiers (Fig. \ref{fig:au101result}). According to the visualization results, both classifiers failed to focus on the relevant regions, i.e., the inner brows, both for the raw frames and face crops.

Our hope was that if we forced the classifiers to focus only on the eye regions, they could learn something meaningful for the recognition of AU101. The last two columns in Fig. \ref{fig:au101result} show that although the classifiers sometimes still look everywhere in the eye regions, they become able to pay attention to the exact inner brows in some cases. Based on these results, we believe that for the task at hand, it is critical to give pre-defined ROIs as input to the classifiers. 

\begin{figure*}[!t]
    \centering
    \includegraphics[width=\textwidth]{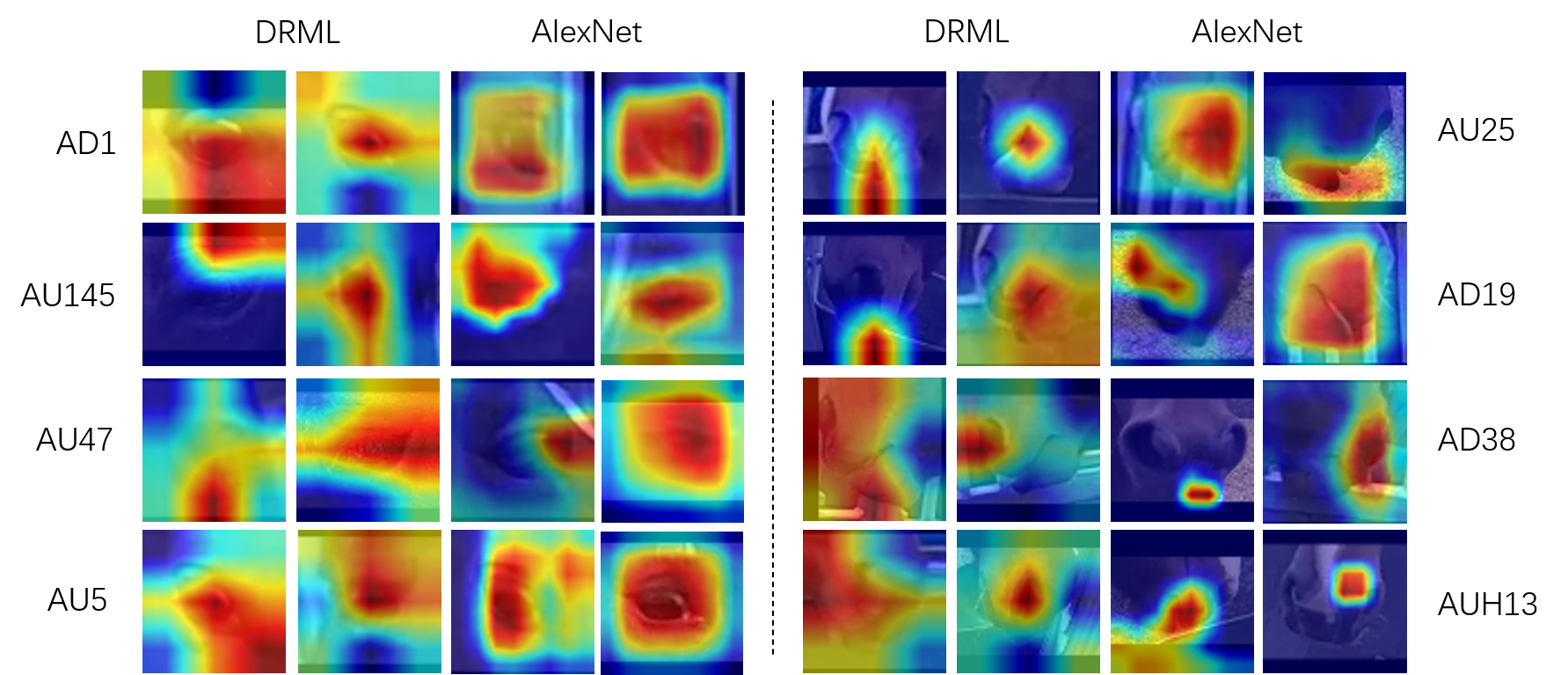}\vspace{-2.5mm}
    \caption{Grad-CAM saliency maps of the different models for binary classification of the other eight AUs. Examples with precise focus are shown on the even columns}
    \label{fig:8ausresult}
\end{figure*}

\subsection{Model Validation on \so{Eight} Other AUs}
Based on the experiments on AU101, we
carried out experiments on \so{eight} other AUs on
the eye and lower face regions
The results are shown in Table \ref{table:otherauresult}.

\begin{table}[!t]
\centering
\caption{Binary Classification for Eight Other AUs. Eye region detection is used for AD1, AU145, AU47, and AU5, while lower face region detection is used for AU25, AD19, AD38, and AUH13.}
\vspace{-1.5mm}
\scalebox{0.9}{
\begin{tabular}{|c|l|l|l|l|}
\hline
\multirow{2}{*}{AU} & \multicolumn{2}{c|}{DRML}                                     & \multicolumn{2}{c|}{AlexNet}                                  \\ \cline{2-5}
                                                                      &      \multicolumn{1}{c|}{Accuracy} & \multicolumn{1}{c|}{F1-score} & \multicolumn{1}{c|}{Accuracy} & \multicolumn{1}{c|}{F1-score} \\ \hline
                                                                       AD1                 & 65.2$\pm$6.0                      & 64.0$\pm$10.4                     & 64.1$\pm$6.2                      & 60.8$\pm$8.2                      \\ \hline 
                                                                       AU145               & 57.5$\pm$4.2                      & 57.8$\pm$5.1                      & 59.9$\pm$2.4                      & 57.2$\pm$7.9                      \\ \hline
                                                                       AU47                & 49.6$\pm$3.7                      & 50.7$\pm$9.1                      & 49.4$\pm$2.3                      & 47.9$\pm$8.3                      \\ \hline 
                                                                       AU5                 & 60.5$\pm$8.1                      & 57.6$\pm$8.9                      & 57.9$\pm$8.8                     & 56.9$\pm$12.6                \\ \hline
 AU25                & 59.8$\pm$6.7                      & 57.8$\pm$9.6                      & 63.6$\pm$8.2                      & 57.9$\pm$12.6                     \\ \hline 
                                                                       AD19                & 64.6$\pm$5.0                      & 59.6$\pm$8.9                      & 61.8$\pm$5.0                      & 58.0$\pm$8.1                      \\ \hline 
                                                                       AD38                & 58.5$\pm$4.1                      & 57.3$\pm$6.4                      & 60.9$\pm$7.1                      & 57.7$\pm$10.1                     \\ \hline
                                                                       AUH13               & 58.6$\pm$2.7                      & 53.2$\pm$6.4                      & 60.0$\pm$4.3                      & 56.1$\pm$9.1                      \\ \hline
\end{tabular}}
\label{table:otherauresult}
\end{table}

In these experiments, generally, \so{the difference is not large} between the performance of DRML and AlexNet, but the DRML typically \so{showed a more stable performance across the different subject folds} than AlexNet. For most AUs, the results lie close
to the those on AU101, except for AU47 (half blink). Moreover, AU47 is sometimes confused with AU145 (blink). We believe that this is because the difference between the presence or absence of AU47 is too small in still images. Our framework would need to be extended to take sequences of images as input to 
to detect it, as in e.g. \cite{broome2019dynamics}.

\so{Similarly, we note} that theoretically, we cannot distinguish AU145 (blink) from AU143 (eye closure) in still images because the sole difference between these is the duration of the time the eyes remain closed. However, since
\so{the AU143 class has too few \so{samples} in our dataset,} we did not include
\so{it} in our experiment\so{s}.
Therefore, the bias of our dataset causes this "good" result of AU145.

To further validate our models, we
visualized their saliency maps\so{,} shown in Fig. \ref{fig:8ausresult}. Similar to AU101, the classifiers are in many cases able to pay attention to the correct regions, such as eyelid, nostril, corner of mouth, and tongue (the even columns in Fig. \ref{fig:8ausresult}), if we crop the pre-defined related regions out before training for classification.

\section{Conclusions}

In this project, we proposed a framework for automated detection of equine facial AUs and descriptors 
and showed that our framework could help the classifiers to focus on more relevant regions and to improve the classification accuracy.

There are many avenues to explore in the future. Firstly, because the dataset used in this article is quite small and unbalanced, deeper models such as VGG and ResNet cannot be trained well, and multi-label learning is not suitable. These techniques will be explored when we collect enough data. Secondly, we are aware that the attention of our model is not fully stable, and we would like to add an attention mechanism to the classification models to make our framework more effective. Finally, our framework currently does not work well for the EADs. This is probably due to the many possible positions of the ears, which are extremely mobile and rarely still in horses. EADs are therefore probably best determined from video. This is also the case for the blinking AUs (AU47 and AU145). A future direction is therefore to extend the method to the temporal domain.



\section*{Acknowledgments}

The authors would like to thank Elin Hernlund, Katrina Ask, and Maheen Rashid for valuable discussions. This work has been funded by Vetenskapsr{\aa}det and FORMAS.

\balance
{\small
\bibliographystyle{ieee}
\bibliography{references}

\begin{thebibliography}{10}\itemsep=-1pt

\bibitem{broome2019dynamics}
S.~Broom{\'e}, K.~B. Gleerup, P.~H. Andersen, and H.~Kjellstr\"om.
\newblock Dynamics are important for the recognition of equine pain in video.
\newblock In {\em IEEE Conference on Computer Vision and Pattern Recognition},
  2019.

\bibitem{cohn2007observer}
J.~F. Cohn, Z.~Ambadar, and P.~Ekman.
\newblock Observer-based measurement of facial expression with the {Facial
  Action Coding System}.
\newblock {\em The Handbook of Emotion Elicitation and Assessment},
  1(3):203--221, 2007.

\bibitem{diogo2009face}
R.~Diogo, B.~A. Wood, M.~A. Aziz, and A.~Burrows.
\newblock On the origin, homologies and evolution of primate facial muscles,
  with a particular focus on hominoids and a suggested unifying nomenclature
  for the facial muscles of the {Mammalia}.
\newblock {\em J.~Anatomy}, 215, 2009.

\bibitem{ekman1993emotion}
P.~Ekman.
\newblock Facial expression and emotion.
\newblock {\em American Psychologist}, 48(4):384--392, 1993.

\bibitem{ekman1978facial}
P.~Ekman and W.~V. Friesen.
\newblock {\em Facial Action Coding System}.
\newblock Consulting Psychologists Press, 1978.

\bibitem{ekman2002facial}
P.~Ekman, W.~V. Friesen, and J.~C. Hager.
\newblock {\em Facial action coding system [E-book]}.
\newblock Research Nexus, 2002.

\bibitem{he2016deep}
K.~He, X.~Zhang, S.~Ren, and J.~Sun.
\newblock Deep residual learning for image recognition.
\newblock In {\em IEEE Conference on Computer Vision and Pattern Recognition},
  2016.

\bibitem{hummel2020automatic}
H.~I. Hummel, F.~Pessanha, A.~A. Salah, T.~{van Loon}, and R.~C. Veltkamp.
\newblock Automatic pain detection on horse and donkey faces.
\newblock In {\em IEEE International Conf.~Automatic Face and Gesture
  Recognition}, 2020.

\bibitem{descovich2017face}
{K.~A.~Descovich, J.~Wathan, M.~C.~Leach, H.~M.~Buchanan-Smith, P.~Flecknell.,
  et al.}
\newblock Facial expression: An under-utilized tool for the assessment of
  welfare in mammals.
\newblock {\em ALTEX}, 34, 2017.

\bibitem{krizhevsky2012imagenet}
A.~Krizhevsky, I.~Sutskever, and G.~E. Hinton.
\newblock Imagenet classification with deep convolutional neural networks.
\newblock In {\em Neural Information Processing Syst.}, 2012.

\bibitem{Liu_2016}
W.~Liu, D.~Anguelov, D.~Erhan, C.~Szegedy, S.~Reed, C.-Y. Fu, and A.~C. Berg.
\newblock {SSD}: Single shot multibox detector.
\newblock In {\em European Conf.~Computer Vision}, 2016.

\bibitem{lu2017estimating}
Y.~Lu, M.~Mahmoud, and P.~Robinson.
\newblock Estimating sheep pain level using facial action unit detection.
\newblock In {\em IEEE Int.~Conf.~Automatic Face and Gesture Rec.}, 2017.

\bibitem{Pessanha2020}
F.~Pessanha, K.~McLennan, and M.~Mahmoud.
\newblock Towards automatic monitoring of disease progression in sheep: A
  hierarchical model for sheep facial expressions analysis from video.
\newblock In {\em 2020 15th IEEE International Conference on Automatic Face and
  Gesture Recognition (FG 2020)}, pages 387--393, 2020.

\bibitem{rashid2020equifacs}
M.~Rashid, K.~B. Gleerup, A.~Silventoinen, and P.~H. Andersen.
\newblock Equine facial action coding system for determination of pain-related
  facial responses in videos of horses.
\newblock {\em PLOS ONE}, accepted, 2020.

\bibitem{Redmon2018YOLOv3}
J.~Redmon and A.~Farhadi.
\newblock {\em YOLOv3: An Incremental Improvement}.
\newblock arXiv preprint arXiv:1804.02767, 2018.

\bibitem{NIPS2015_5638}
S.~Ren, K.~He, R.~Girshick, and J.~Sun.
\newblock Faster r-cnn: Towards real-time object detection with region proposal
  networks.
\newblock In {\em Neural Information Proc.~Syst.} 2015.

\bibitem{Selvaraju_2019}
R.~R. Selvaraju, M.~Cogswell, A.~Das, R.~Vedantam, D.~Parikh, and D.~Batra.
\newblock Grad-cam: Visual explanations from deep networks via gradient-based
  localization.
\newblock {\em International Journal of Computer Vision}, 128(2), 2019.

\bibitem{shao2019facial}
Z.~Shao, Z.~Liu, J.~Cai, Y.~Wu, and L.~Ma.
\newblock Facial action unit detection using attention and relation learning.
\newblock {\em IEEE Transactions on Affective Computing}, 2019.

\bibitem{simonyan2014very}
K.~Simonyan and A.~Zisserman.
\newblock Very deep convolutional networks for large-scale image recognition.
\newblock In {\em International Conf.~Learning Representations}, 2015.

\bibitem{wathan2015equifacs}
J.~Wathan, A.~M. Burrows, B.~M. Waller, and K.~{McComb}.
\newblock {EquiFACS}: the equine facial action coding system.
\newblock {\em PLOS ONE}, 10(8), 2015.

\bibitem{williams2002pain}
A.~C. Williams.
\newblock Facial expression of pain: An evolutionary account.
\newblock {\em Behav.~Brain Sciences}, 25, 2002.

\bibitem{zhao2016deep}
K.~Zhao, W.-S. Chu, and H.~Zhang.
\newblock Deep region and multi-label learning for facial action unit
  detection.
\newblock In {\em IEEE Conf.~Computer Vision and Pattern Rec.}, 2016.

\end{thebibliography}
}

\end{document}